# Helping university students to choose elective courses by using a Hybrid Multi-criteria Recommendation System with Genetic Optimization

A. Esteban, A. Zafra*, C. Romero

*University of Cordoba, Dept. of Computer Science, 14071, Cordoba, Spain*

## Abstract

The wide availability of specific courses together with the flexibility of academic plans in university studies reveal the importance of Recommendation Systems (RSs) in this area. These systems appear as tools that help students to choose courses that suit to their personal interests and their academic performance.

This paper presents a hybrid RS that combines Collaborative Filtering (CF) and Content-based Filtering (CBF) using multiple criteria related both to stu-dent and course information to recommend the most suitable courses to the students. A Genetic Algorithm (GA) has been developed to automatically dis- cover the optimal RS configuration which include both the most relevant criteria and the configuration of the rest of parameters. The experimental study hasused real information of Computer Science Degree of University of Cordoba (Spain) including information gathered from students during three academic years, counting on 2500 entries of 95 students and 63 courses. Experimental results show a study of the most relevant criteria for the course recommendation, the importance of using a hybrid model that combines both student informa-tion and course information to increase the reliability of the recommendations as well as an excellent performance compared to previous models.

*Corresponding author

*Email addresses:* aestebant@uco.es (A. Esteban), azafra@uco.es (A. Zafra), cromero@uco.es (C. Romero)

# 1. Introduction

University studies usually involve in their curriculum various elective courses. These courses have to be chosen by the students between many options, being essentials to finalize the studies and to obtain their university degree. In this scenario, students spent so much time searching for information about the available different courses in order to make the best decision regarding their academic plan. Making this decision may not be trivial since students do not have enough information. Thus, generally, they are influenced by other college students' comments. In this context, it is important to have into account the preferences and interests of students: some students look at specific contents, others prefer easy courses or obtain high grades and others could look for particular professors with which they have had good experience in the past. With these conditions, it is very important to consider and analyze which one of the different opinions that a student can receive is relevant to his or her interests. A way to resolve this problem is by means of Recommendation Systems (RS) that help students to make a good decision adapted to their specific preferences. Some examples that show the relevance of these systems can be found in [1].

The main goal of RSs is to deliver customized information to a great variety of users according to their preferences. The most common RSs are the Collaborative Filtering (CF) and the Content-based Filtering (CBF) [2]. CF recommends items based on ratings of similar users, while CBF recommends items based on content of similar items to the user profile. In the last years, hybrid techniques as well as multiple criteria approaches have gained importance, since they help to solve the problems that each basic technique has separately.

In the literature, it can be found different hybrid RSs [3] and the use of multiple criteria [4] that show the relevance of these techniques in educational data mining. Focusing our attention on course recommendation, previous works do not include many criteria neither they study the real influence of each criterion in the recommendation. In this paper, the main contributions can be summarized as follows:

- A proposal of a hybrid multi-criteria course recommendation system that combines CF using criteria related to students' information, such as rat- ings, grades and branches, in conjunction with CBF based on criteria related to course information, such as competences, professors, theoretical and practical contents and knowledge area. An extensive number of criteria are used compared to previous proposals.

- A proposal of Genetic Algorithm (GA) which automatically optimizes both the weights assigned to each criterion and other configuration parameters used in the RS, such as the similarity measures for each criterion and the size of neighborhood. This algorithm is applied previously as external step of the SRs. Thus, firstly the GA is trained to produce theoptimized configuration for the RS, then the RS model is built with this configuration. Finally, RS produces the recommendations for the students with the guarantee that the best possible configuration is being used. A specific GA is designed and adapted to course recommendation. An extensive number of elements are configured by GA compared to previous proposals.

- Different studies are carried out to show the advantages of this proposal. First, an analysis of the relevance of different criteria to determine the most relevant. A similar study is not included in previous multiple criteria works. Second, a study about the importance of automatically to optimize the RSs is also carried out. Thus, our proposal is compared to different versions considering different criteria and CF and CBF independently.Third, a comparison with previous works is considered in the experimental study. Finally, it is shown a study case to show the recommendation given by this system to a specific user.

The experimental study uses information of Computer Science Degree of University of Cordoba (Spain). It is included information about ratings and grades that has been collected of students during three academic years (2016-2018). In total, it has been collected 2500 ratings, and their related grades, belonging to 95 students and 63 courses. For the evaluation of the RS accuracy, it has been implemented a stratified cross-validation process keeping a balance between the number of ratings received per course across the different parti- tions. Experimental results highlight the importance of using several criteria in the recommendation process. Further, our tests show that combining CF and CBF techniques in a hybrid RS approach improves the results of each one separately. Concerning to the relevance that each criterion has, in general student information seems to be more useful than the course, being the ratings of simi- lar students, the most important criterion. With respect to course criteria, the most important factor for recommending a course to a student is the coincidence of professors with courses he or she liked in the past.

The remainder of this paper is organized as follows. In Section 2 a reviewof the work related to our proposal is described. In Section 3 it is presented and specified the proposed methodology. Section 4 shows the experimental study carried out. Finally, conclusions obtained and future work are provided in Section 5.

## 2. Related work

In recent years, Educational Data Mining (EDM) has become an established research field due to the expansion of information systems that support the learning process in any educational scope. Romero and Ventura [5] present an updated survey of most important studies in the field. Equally, a complete analysis of most important techniques in recent years of EDM can be found in the survey of Peña-Ayala [6]. In parallel, RSs have been emerged as an useful technique to guide to the users in different domains where there is a vast amount of information available, such e-commerce, music or movies [7]. Within this field, the most common technique is the CF, based in similar users, follows by CBF, based in similar items. Moreover, hybrid techniques is increasing their use because they take advantages of different models. A very successful application of EDM techniques is the development of Educational RS. One of the firsts applications in the field can be found in [8], that explore the extraction of students' learning requirements and use matching rules to generate personalized recommendations of learning activities in a context of e-learning environments. Since then, these systems have been applied to a broad domain, ranging from automatic suggestions for the assignment of courses timetables and classrooms [9], to recommendations for creation of a long-term course planning that takeinto account constraints concerning to both student and courses [10]. In this context, RSs have been thoroughly applied to the problem of course recom- mendation from different approaches. Recently, Iatrellis et al. [1] present a systematic review of most recent RS applied to course selection from an exper- imental perspective encompassed in the Academic Advising Systems discipline.

### *2.1. Recommendation Systems using Genetic Algorithms*

It is important to notice that the combination of RS with GA has not been barely explored in the field of course recommendation. Extending the study to the recommendation in other contexts, any references can be found. Although the works show a good performance, the number of references found is very reduced:

- Linqi and Congodon [11] introduce a hybrid RS centered in e-commerce where GA is used to combine the RS output. Concretely, the outputs are aggregated in a linear combination with specific weights that are optimized by a GA.

- Hwang [12] proposes a multi-criteria CF to recommend movies. The proposal treats the multi-criteria recommendations as optimization problems and applies a weighted average method by combining values from differ- ent criteria. Concretely, GA is used for optimal feature weighting. This proposal uses GA as part of the RS, so it has to be executed for each recommendation carried out by RS.

- Bobabilla et. al. [13] propose a similarity measure between users ratings that it is applied to recommendation of movies. The GA is used to find the optimal weights of this similarity function formulated via a simple linear combination of values and weights.

- Salehi et al. [14] propose a hybrid RS for the recommendation of learning materials in Moodle. The proposal uses explicit attributes based on the ratings given by the students and implicit attributes that use a GA to obtain specific weights for each student.

Our proposal of GA tries to combine the different ideas proposed in previous works. Thus, it is designed a GA where the representation of solutions and genetic operators are adapted to course recommendation problem. The main features of our proposal are the followings:

- The GA is applied as a prior stage of RS. Thus, GA optimizes with train- ing data the parameter configuration of RS. Then, the RS is configured according to these parameters and the specific recommendations to users can be carried out. The idea is that the GA does not introduce morecomputation time in each recommendation provided.

- The GA considers the optimization of weights for each criterion both CF and CBF system. Thus, each criterion will have a specific relevance to determine the final recommendation.

- The GA considers the optimization of the similarity measures. Thus, each criterion can use different similarity measures and the GA will optimize the most appropriate for each one of them.

- The GA considers the optimization of size of neighborhood. Thus, the size of neighborhood will be optimized to the most appropriate value.

- Finally, The GA considers the optimization the outputs of the hybrid system. Thus, the GA will determine a weight to obtain the relevance of each system used in our hybrid RS. Concretely, CF and CBF systems.

In conclusion, the GA has been designed to have into account all possible configurations of our RS that it has been designed to consider an extensive number of criteria and similarity measures. For it, individuals and genetic operators have been designed according to these particularities, they are specified in section 3.3.

*2.2. Recommendation system for course recommendation*

Nowadays, students have many options when they want to take a course. Usually, it is complicated for students take that decision. In this context, recommendation systems have been proposed as tools that help students to make their choice. In this section, it is carried out a review on CF, CBF and hy-brid techniques applied to courses recommendation. Systems based on CF are widely used. Chang et al. [17] presented a two-stage user-based CF process using an artificial immune system (AIS) for the prediction of student grades. In order to address the the problem of the amount of feedback required from students to produce recommendations, authors segregated the students' popu- lation with demographic information and they introduced a control mechanism that filters courses whose instructors have a low rating. Taha [15] introduced an XML user-based collaborative system which advises a student to take courses that were taken successfully by students with the same interests and academic performance. The students' categorization is based on course features such as memorization skills or programming skills, among others. Bakhshinategh et al. [4] explored the inclusion of a normalized system to describe the competences that a course provides and the courses that helped to the students to achieve them. Ganeshan and Li [16] designed a web-based RS that uses K-means algorithm to determinate the similarity of the students.

With respect to systems based on CBF, recent and relevant proposals can be found. Mostafa et al. [18] presented a case-based reasoning that made recommendations based on matching features associated to courses. Ontology-driven software development [19] is also explored as CBF system. In this case, modeling various aspects of the academic plan in order to recommend courses that help to complete the required credits to the students. Recently, Ma et al. [20] explored the application of semantic similarity to courses description for providing recommendations.

Table 1: Comparison between proposals

| **Algorithms** | **Criteria** | | **Similarity Measure** |
|---|---|---|---|
| | **Student information** | **Course information** | |
| Biclustering with XML-based CF [15] | • Academic skills | | • Cosine similarity |
| User-based CF [4] | • Ratings over course competences | | • Pearson correlation |
| Clustering using CF [16] | • Average grades<br>• Demographic data | | • Taxicab distance |
| AIS with clustering based on CF[17] | • Grades over courses | • Professors | • Cosine similarity<br>• Pearson correlation |
| Case-based reasoning [18] | | • Courses key words | • Cosine similarity |
| Ontology based agent [19] | Course completed | • Synonymous set<br>• Credits of course | • Similarity based on ontology relations |
| Clustering & semantic similarity [20] | | • Description of courses | • Euclidean distance over frequency n-gram vectors |
| Item-based filtering & User-based CF [21] | • Ratings over courses | • Course area<br>• Professors | • Cosine similarity<br>• Pearson correlation |
| Rule-based & case-based reasoning [22] | • Grades over courses | • Course prerequisites | • Percentage of match pairs of students information based on their total pair |
| Clustering & association rules based on CF [3] | • Grades over courses | | • Taxicab distance<br>• Rules support & confidence |
| N-gram query expansion & ontology [23] | | • Courses key words<br>• Synonymous database | • Terms frecuency<br>• Classification based on ontology relations |
| Fuzzy tree matching, knowledge-based filtering & CF [24] | • Requirements for learning categories<br>• Courses done<br>• Previous curriculum | • Learning categories and subcategories<br>• Sequential relations | • Fuzzy tree similarity developed by authors |
| **Our proposal** | • Ratings over courses<br>• Grades over courses<br>• Branch of students | • Professors<br>• Course contents<br>• Knowledge area<br>• Course competences | • Euclidean distance<br>• Taxicab distance<br>• Pearson correlation<br>• Spearman correlation<br>• Jaccard index<br>• Log-likelihood function<br>• Semantic similarity |

Finally, hybrid RSs that combine several techniques of recommendation are taking more and more importance. Unelsrød [21] explored the combination of CF and CBF through the generation of recommendations generated indepen- dently and presented together. That study showed the importance of using an existing and relatively large dataset to test the RS.

Daramola et al. [22] presented a hybrid CBF system that combines asso- ciation rules and case-based reasoning with courses-related information. Al- Badarenah and Alsakran [3] combined CF with association rules in order to predict students performance. Gulzar et al. [23] explored the use of an ontology along with N-gramm queries. Wu et al. [24] proposed a CF combined with fuzzy trees to represent both student and learning activities information.

Table 1 shows a summary of the main characteristics of each proposal. It is considered both student and course specific criteria, as well as, the similarity measures utilized that are a key element in the RSs for finding the students and course more similar. It is relevant to highlight that most proposals use one or two criteria and one or two similarity measures. After this study, it can be summarized that the main contributions of our proposal in relation to the related work:

- Our proposal uses a very representative number of criteria and similarity measure. Concretely, it is used seven different criteria combining both student and course information, as well as, seven different similarity measures.

- All this information is automatically configured by means of GA that determines the most relevant criteria for the recommendation (a depth study of the most relevant criterion and the most appropriate similarity measures for each criterion are analyzed in experimentation section).

## 3. Proposed methodology

Our proposed methodology has several steps (see Figure 1). First, it is addressed the description and processing of the used data. Then, it is detailed the proposed hybrid multi-criteria RS. This system recommends courses to university students based on several criteria related to both student and course information. Finally, it is described the designed optimization method that assigns a weight to each criterion and optimizes the rest of RS parameters automatically. This method allows to identify the relevance of each criterion using a system of weights. Thus, the most relevant criteria have higher weights while the less ones have lower weights. Furthermore, the method finds the optimal configuration for the parameters of the proposed RS, such as, the similarity measures and neighborhood size. Each one of these steps are described in detail in the following section.

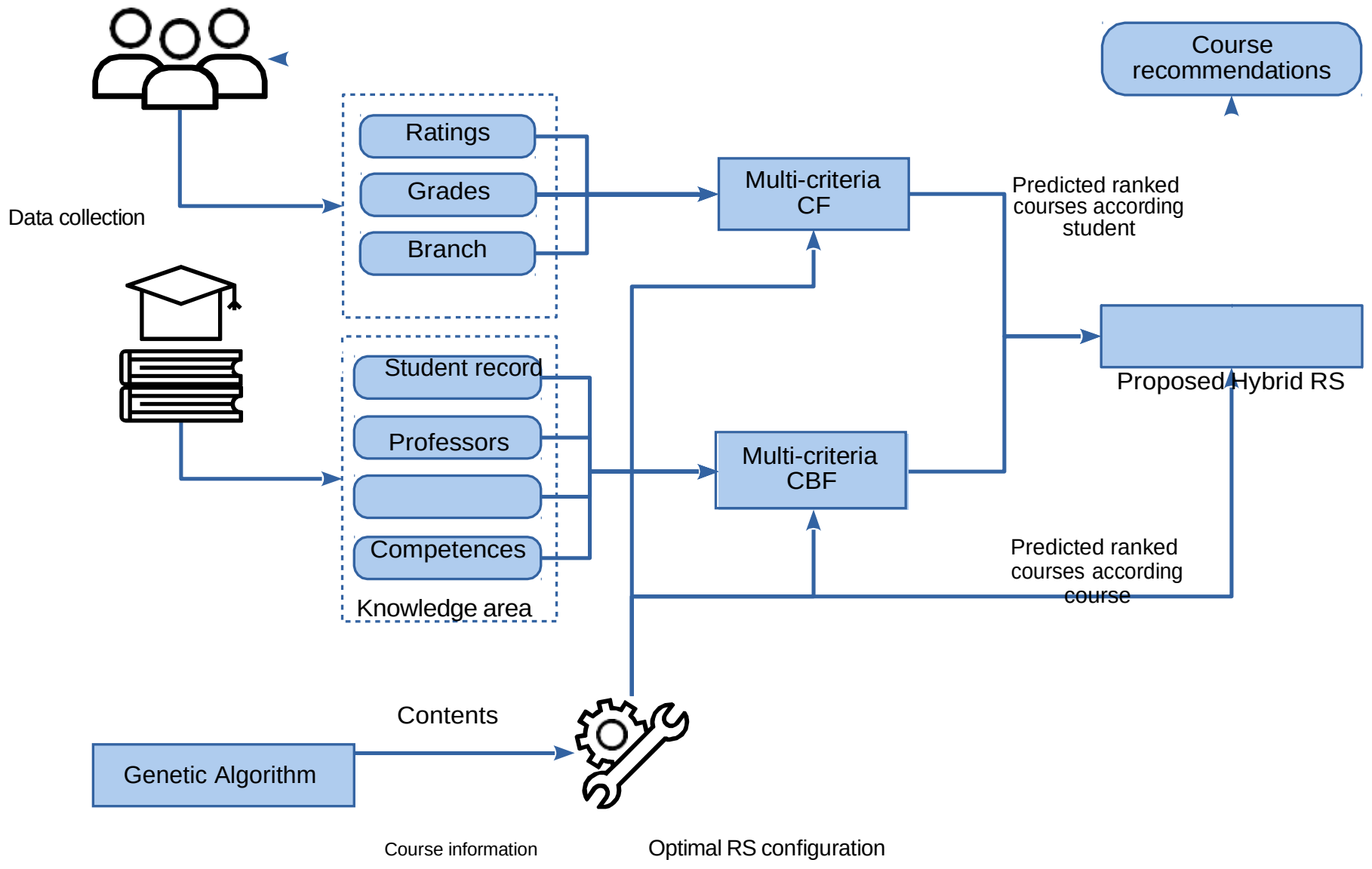

Figure 1: Steps of the proposed hybrid course recommendation system

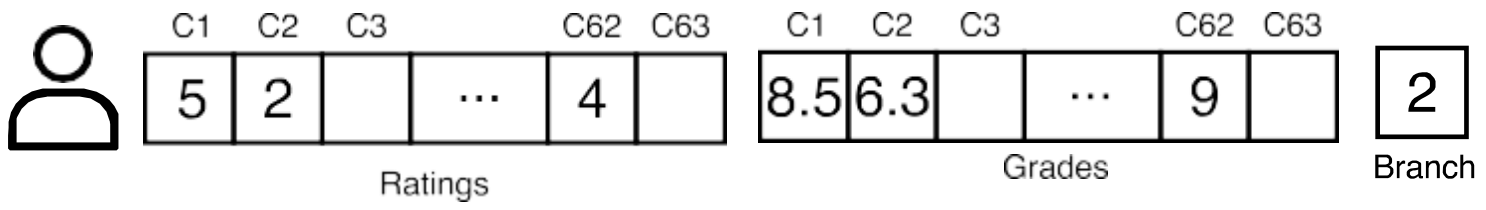

Figure 2: Student information

### 3.1. *Data description and processing*

This work has been developed using student and course information gathered from Computer Science Degree at University of Cordoba, Spain.

#### 3.1.1. *Student information*

Student information has been obtained through surveys carried out during three academic years (from 2016 to 2018). The information includes 95 students and 2500 ratings of 63 courses included in the curriculum. The information considered is the following (see Figure 2):

- A rating of the overall student's satisfaction for each course in a Likert scale of 5 points. Non-taken courses are assigned an empty value.
- The grade obtained by the student in each course. It is a decimal value in the range $[0, 10]$. Non-graded courses are assigned an empty value.
- The branch selected by the student. Concretely, the studied degree offers three branches for specialization: Computation (identified by 1), Com- puter Engineering (identified by 2) or Software Engineering (identified by 3). It is a numeric identifier (from 1 to 3) representing each branch.

#### 3.1.2. *Course information*

There is information about 63 obtained from course catalogue of Computer Science Degree at University of Cordoba. The criteria selected for each course are the following (see Figure 3):

- The professors involved in the course. It is represented as a vector with an index for each professor in the degree. Its value is 1 if the professor teaches in that course or 0 if the professor does not teach.

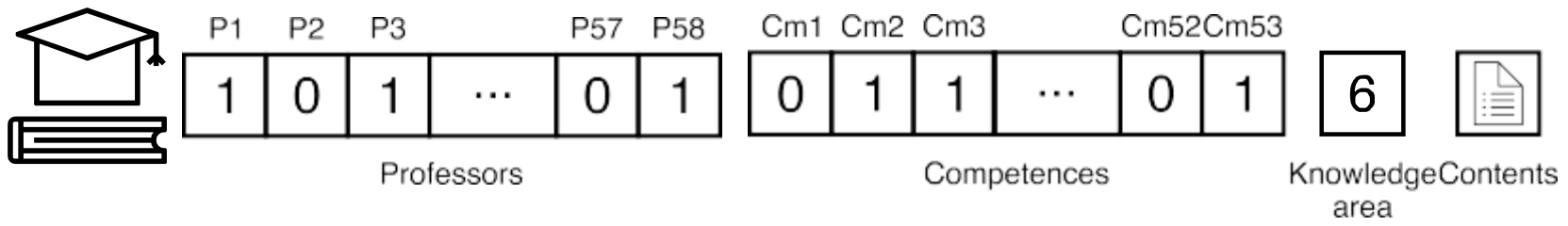

Figure 3: Course information.

- Course competences or skills. It is represented as a vector with an index for each competence in the degree. Its value is 1 if the competence is contained in the course or 0 if the competence is not contained.
- The knowledge area of the course. It is represented as a numeric value. There are eight different knowledge areas in the Computer Science degree (integer value from 1 to 8).
- Theoretical and practical contents of the course. It is represented as a frequency vector of keywords obtained by text mining from teaching guide. The teaching guided is obtained from web page[1].

### *3.2. Multi-criteria Hybrid Recommender System*

The proposed hybrid RS combines two multi-criteria systems. One of them is based on CF model and the other is based on CBF model. Thus, the estimation of the preference $p$ of a student $i$ over a course $j$ will be computed as it is indicated in equation 1.

$$p_{i,j} = \alpha \cdot CF_{i,j} + \beta \cdot CBF_{i,j} \quad (1)$$
$$With\ \alpha + \beta = 1$$

where $CF_{ij}$ is the recommendation given by CF model for student $i$ over the course $j$ and $CBF_{ij}$ is the recommendation given by CBF model for student $i$ over the course $j$.

Both the CF and the CBF system provide the estimation in range $[1, 5]$ and the final preference $p$ also is in that range. The parameters that determine the relevance given to each system considered are $\alpha$ and $\beta$ weights which must be configured. In next sections, the details of CF and CBF models are specified.

[1]http://www.uco.es/eps/node/619

### *3.2.1. Collaborative Filtering using student information*

The developed CF system estimates the ratings for new courses from the ratings given to these courses by similar students. The different criteria are used to find the most similar students. Thus, for each pair of students $i$ and $j$, three similarity measures are considered:

- Similarity by ratings $R_{ij}$. This measure computes the ratings introduced by each student and calculates the distance between students using Euclidean or Taxicab metrics or a linear correlation approach with Pearson or Spearman coefficients.
- Similarity by grades $G_{ij}$. This measure computes the grades obtained by each student and calculates the distance between students according to this criterion using Euclidean or Gaxicab metrics or following a correlation approach with Pearson or Spearman coefficients.
- Similarity by branch $B_{ij}$. This measure checks if the branch chosen by each student is the same or it is not.

Finally, these criteria are aggregated in a linear combination that produces a global similarity value for each pair of students $i$ and $j$ (see equation 2):

$$s_{ij} = \alpha \cdot R_{ij} + \beta \cdot G_{ij} + \gamma \cdot B_{ij} \quad (2)$$

$$With\ \alpha + \beta + \gamma = 1$$

This measure is very flexible because it can be configured assigning different relevance to each criterion, so that, it is necessary to configure the weights $\alpha$, $\beta$ and $\gamma$ that determinate the importance given to each criterion.

Moreover, student-based CF generates a neighborhood with the most similar students according to the similarity measure shown in equation 2 to carry out the recommendation. Therefore, the size of this neighborhood also need to be configured in our system.

*3.2.2. Content-based Filtering with course information*

The developed CBF system recommends courses to students based on his or her own previous ratings of the most similar courses. The different criteria considered are used to find the most similar courses. Thus, for each pair of courses $i$ and $j$, four similarity measures are combined:

- Similarity considering professors, $P_{ij}$. This measure computes the professors that teach in each course and calculates the similarity based on how many professors have in common each course. Similarity can be computed following a set theory approach with Jaccard index or a probabilistic approach with log-likelihood function.
- Similarity considering competences, $Cm_{ij}$. This measure computes the common competences in each course. Similarly, it can be calculated using Jaccard index or log-likelihood function.
- Similarity considering knowledge area, $S_{ij}$. This measure checks if the knowledge area of each course is the same or it is not.
- Similarity considering contents, $Cn_{ij}$. This measure applies text-mining over theoretical and practical contents of each course. The following steps are taken to obtain a similarity coefficient:

  1. Indexing the *Contents* specified in the courses' teaching guides: we have implemented a custom text parser based on the language (in our case, Spanish). It is used in conjunction to a set of stop words adapted to the domain. As a result, a list of tokens is obtained along with their frequency for each document.
  2. For each pair of courses, $i$, $j$, a set $B$ is created as the union of the tokens of both courses. For each course, a vector $\vec{i}$ and $\vec{j}$ is built with so many elements as there are in $B$. These vectors contain the frequency of each token. Finally, each vector is normalized using the *l1* norm. Thus, it is obtained the relative frequencies to each pair of courses.

3. Cosine similarity is applied on frequency vectors obtaining the similarity measure, $Cn_{ij}$. This measure is integrated in the course global similarity:

$$cos(\vartheta) = \frac{\vec{i} \cdot \vec{j}}{\|\vec{i}\| \cdot \|\vec{j}\|} = \frac{\sum_{k=1}^{n} i_k j_k}{\sqrt{\sum_{k=1}^{n} i_k^2} \sqrt{\sum_{k=1}^{n} j_k^2}} \tag{3}$$

Finally, these criteria are aggregated in a linear combination to obtain the similarity value for each pair of courses $i$ and $j$ (see equation 4).

$$s_{ij} = \alpha \cdot P_{ij} + \beta \cdot Cm_{ij} + \gamma \cdot S_{ij} + \delta \cdot Cn_{ij} \tag{4}$$
$$With\ \alpha + \beta + \gamma + \delta = 1$$

Our similarity measure is highly flexible because it is able to indicate different relevance to each factor to obtain the most similar courses. Thus, it is necessary to configure the weights $\alpha$, $\beta$, $\gamma$ and $\delta$ that determinate the importance givento each criterion.

### 3.3. Optimization with genetic algorithm

The proposed RS has multiple criteria that must be pondered with a weight in order to indicate their relevance in the recommendations. Further, there are RS parameters to configure, such as, the metrics to compute the similarity between students or courses. Thus, we propose a GA that automatically finds an optimal configuration for the RS. We have developed a variation of the CHC algorithm developed by Eshelman [25]. This algorithm follows an adaptive search approach adapted to the specific characteristics of this problem. CHC isa classic algorithm that has shown an excellent performance in similar scenarios [26, 27]. It combines high diversity given by incest prevention where offspring are obtained from different parents and the population is restarted when it is stagnant. Moreover, this algorithm gets an elevate convergence, given bya elitist selection that preserves the best individuals in each generation. The architecture overview of the GA adapted for our problem can be view in Figure

4. Its ain characteristics are addressed in following sections.

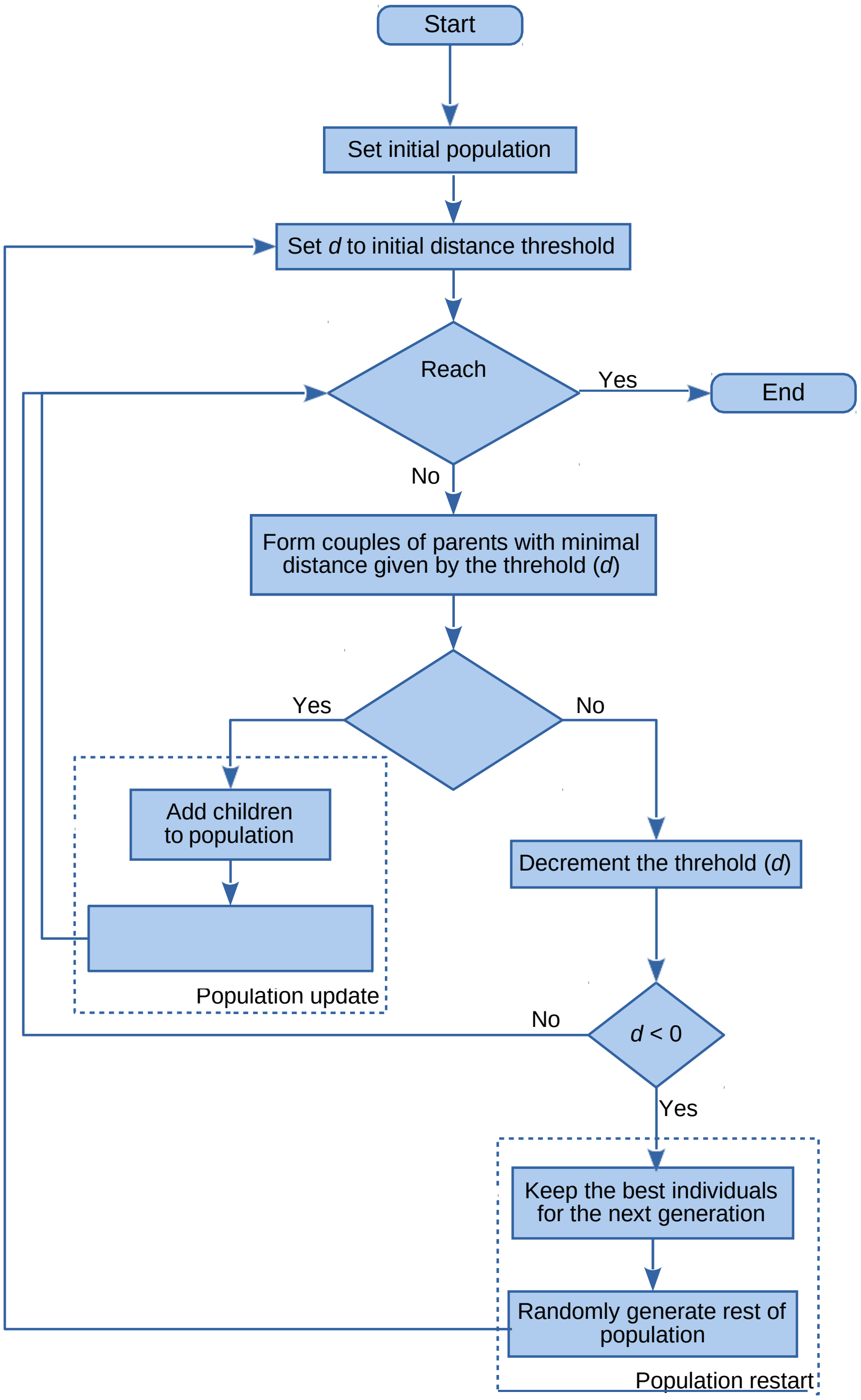

Figure 4: Flowchart of the proposed GA

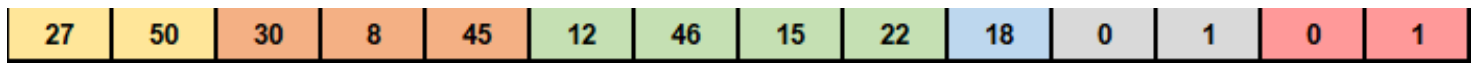

(a) Individual Genotype

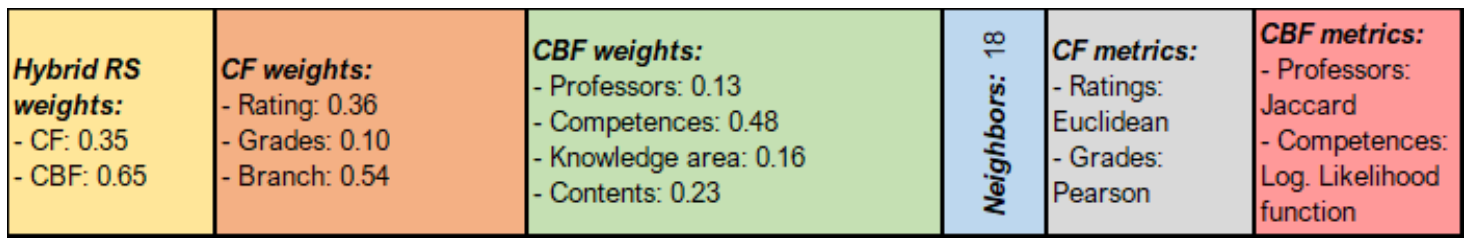

(b) Individual Phenotype

Figure 5: Individual representation in the proposed GA

### 3.3.1. *Representation of individuals*

The goal of this GA is to find an optimal configuration for our multi-criteria hybrid RS, so the individuals are possible configurations of our system. These configurations are represented as a chromosome consisting of 14 genes coded as integers grouped by five categories. On one hand, Figure 5a shows the genotype which represents each individual. It is composed by different integer values that are optimized by our GA. On the other hand, Figure 5b shows the correspond- ing phenotype. It is composed by specific meaning in the configuration of our system. Following, each group of genes is described:

- The first two genes represent the weights used for combine the recommendations given by CF and CBF system, respectively. Specifically, they are the values of $\alpha$ and $\beta$ in Equation 1. They can take values in a range of integer values set in the algorithm configuration. However, at the evaluation moment, a normalization process transforms the $x_i$ integer value of these genes to the $z_i$ decimal value following the next equation: $z_i = x_i / \sum_{j=1}^{2} x_j$. Thus, it is scaled the value of these genes to $[0, 1]$ range. Then, the $z_i$ value is rounded to two decimal places and it is ensured that the sum of all values is 1. According to the configuration given by the example of Figure 5: $\alpha = z_1 = 27/(27 + 50) = 0.35$ and $\beta = z_2 = 50/(27 + 50) = 0.65$.

- The following three genes represent the weights given to each considered criterion attending to student information. They are used to compute similarities in CF system based on students. Specifically, they represents the values of $\alpha$, $\beta$ and $\gamma$ in Equation 2. They can be specified in a range of integer values set in the algorithm configuration. However, at the evaluation moment, a normalization process transforms the $x_i$ integer value of these genes to the $z_i$ decimal value following the next equation: $z_i = x_i / \sum_{j=3}^{5} x_j$. Thus, it is scaled the value of these genes to $[0, 1]$ range. Then, the $z_i$ value is rounded to two decimal places and it is ensured that the sum of all values is 1. According to the configuration given by the example of Figure 5: $\alpha = z_3 = 30/(30 + 8 + 45) = 0.36$, $\beta = z_4 = 30/(30 + 8 + 45) = 0.10$ and $\gamma = z_5 = 45/(30 + 8 + 45) = 0.54$.

- The following four genes represent the weights given to each considered criterion attending to course information. They are used to compute similarities in CBF system based on courses. Specifically, they are the values $\alpha$, $\beta$, $\gamma$ and $\delta$ in Equation 4. They can be specified in a range of integer values set in the algorithm configuration. However, at the evaluation moment, a normalization process transforms the $x_i$ integer value of these genes to the $z_i$ decimal value following the next equation: $z_i = x_i / \sum_{j=6}^{9} x_j$. Thus, it is scaled the value of these genes to $[0, 1]$ range. Then, the $z_i$ value is rounded to two decimal places and it is ensured that the sumof all values is 1. According to the configuration given by the exam- ple of Figure 5: $\alpha = z_6 = 12/(12 + 46 + 15 + 22) = 0.13$, $\beta = z_7 = 46/(12 + 46 + 15 + 22) = 0.48$, $\gamma = z_8 = 15/(12 + 46 + 15 + 22) = 0.16$ and $\delta = z_9 = 22/(12 + 46 + 15 + 22) = 0.23$.

- The next gene corresponds with the size of neighborhood used in CF system based on students. According to the configuration given by the example of Figure 5, the size of neighborhood is 18.

- The following two genes represent the metrics used to compute similarities in CF system based on students, specifically using ratings and gradescriteria. Those genes follow a categorical approach using a numerical identifier for each of the four metric mentioned in Section 3.2.1. According to the configuration given by the example of Figure 5: it is used the*Euclidean distance* to measure the similarity between students' ratings and *Pearson coefficient* to measure the similarity between students' grades.

- The last two genes represent the metrics used to compute similarities in CBF system based on courses, specifically using professors and compe- tences criteria. Those genes follow a categorical or boolean approach us- ing a numerical identifier for each of the two metric mentioned in Section 3.2.2. According to the configuration given by the example of Figure 5: it is used the *Jaccard coefficient* to measure the similarity between professors in each course and *Logarithmic likelihood function* to measure the similarity between competences in each course.

### *3.3.2. Genetic operators*

Different genetic operators are designed for this algorithm. They are described in this section.

*Incest prevention.* This operator promotes the exploration and reduces the genetic drift. Thus, the operator maintains a distance threshold ($d$) that allows to have offspring only from the individuals that are sufficiently distant (incest prevention). This threshold allows to control when it is moment to restart the population because individuals are too similar.

The procedure followed to carry out the incest prevention, it is described in flowchart of Figure 4:

- First, the threshold $d$ is set to an initial value at the start of algorithm. Concretely, it is set to $L/4$, being $L$ the length of the chromosome. This is the value recommended by [25].

- In each generation, couples of individuals are randomly formed to carry out the crossover operator:

  - if no couple of individuals can be crossover because the distance of all couples of individuals is lower than the incest threshold, the threshold is decremented by one unit for the next generation. Thus, in the next generation, the crossover operator will be less restrictive.

 - if at least one couple of individual has a distance between them higher or equal than the incest threshold, threshold is maintained with the same value for the next generation.

- At the end of each generation, it is checked the value of incest threshold. If the incest threshold reaches zero value, it is considered that the population is too similar. Thus, at this point, the algorithm executes the restarting process (it is explained later in this section) and sets the threshold to its initial value.

*Individuals dissimilarity.* The incest prevention is given by the threshold commented previously. In order to maximize the diversification probabilities of diversification of the population, it is necessary to use an appropriate distance measure that identifies dependable the similarity between individuals.

In our problem, the designed chromosome to represent individuals needs to use a personalized distance to avoid introducing noise in the measurement. We propose to use a distance based on the Hamming distance [28]. Thus, the distance between two individuals is the number of genes that have different values between their chromosomes. The specific particularities considered in this metric are shown in Figure 6. Firstly, the genes associated to weights are normalized and they are treated by groups (see Figure 5). Thus, each group of genes has to sum a value of 1. For example, given two individuals $i$ and $j$ whose group of genes has been normalized and set the restrictions so that they sum 1 (see Figure 6), the evaluation of the distance measure is the following:

- When the first group is compared, which it is composed by 2 genes, only two states are possible: the two genes are equal or the two are different. Thus, this group contributes with 0 or 1 to the final similarity distance$d(i, j)$. According to the example of Figure 6, both genes are different, therefore it is added 1 to the distance.

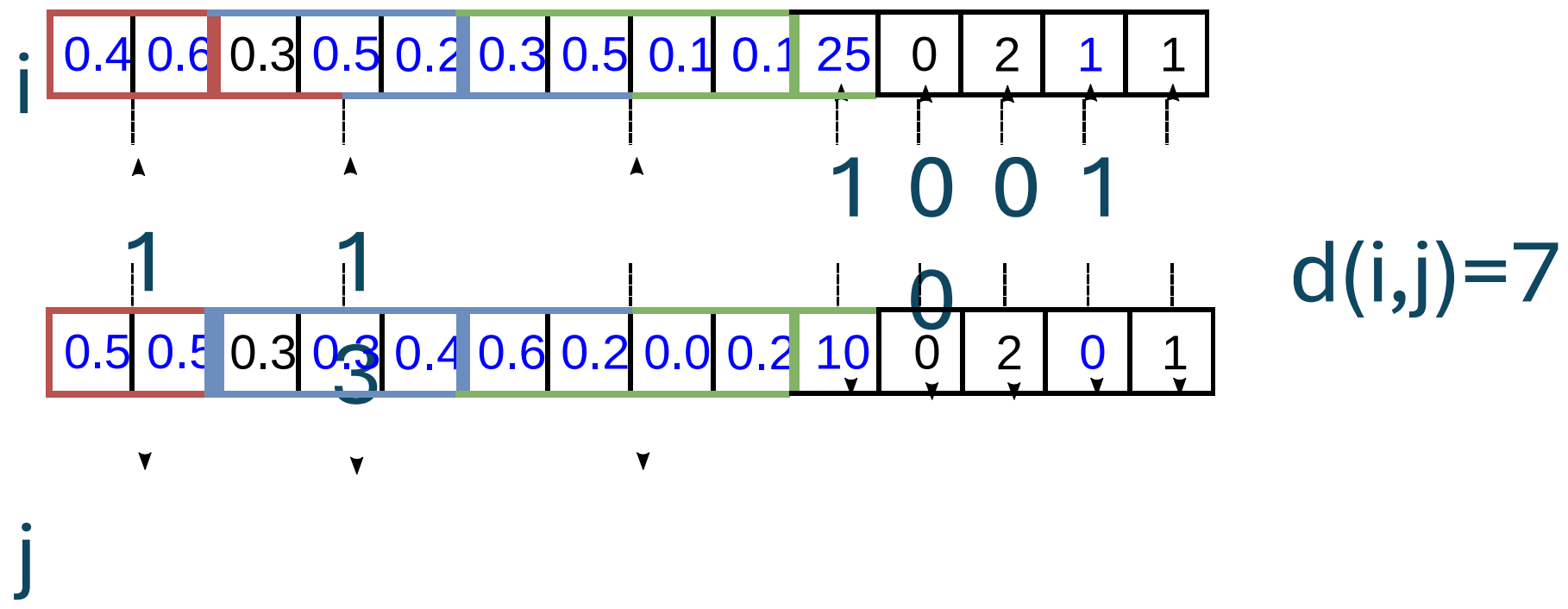

Figure 6: Hamming distance example

- by 3 genes, thereare three possible states: zero, two or three different genes, so this group

  contributes with 0, 1 or 2 to the similarity distance, $d(i, j)$. According to the example of Figure 6, there is one equal gene and the rest is different, therefore it is added 1 to the distance. This it is evaluated in this manner because if two genes are equals, due to the restriction that the sum of the three genes must be 1, the third gene would be always equal.

- When the third group is compared, which it is composed by 4 genes, it is followed the same principle contributing with 0, 1, 2 or 3 to the similarity distance, $d(i, j)$. According to the example of Figure 6, all genes are different, therefore it is added 3 to the distance.

- Finally, the rest of genes are computed individually, if they are equal, they add 0 and otherwise, they add 1. According to the example of Figure 6, it is added 2 to the distance.

*Crossover operator.* It is a uniform-type crossover that takes two parents and generates two children. This crossover evaluates all chromosome positions and for each one randomly assigns each gene of parent to gene of a children.

We propose an adaptation of this crossover for our specific domain in order to deal with the genes that are related are (see Figure 5): (1,2), (3,4,5) and (6,7,8,9).

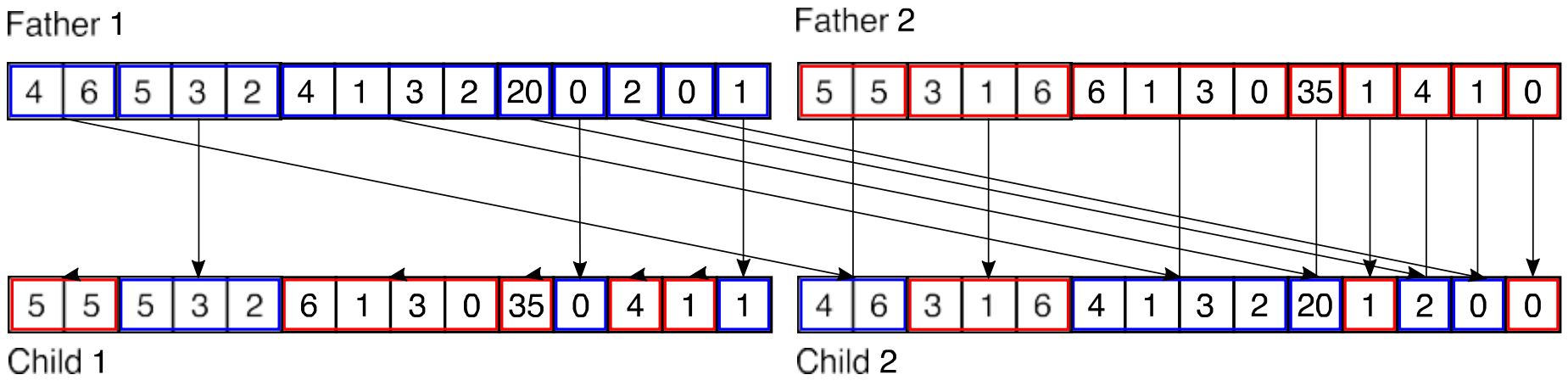

Figure 7: Uniform crossover example

be viewed in Figure 7.

*Updating process.* The population is updated from a generation to the next one by means of an elitism strategy. Concretely, it is united parents and offspring and the individuals that go through the next generation are the best individuals from united population.

*Restarting process.* The chosen process of population updating can introduce a high selective pressure. With the purpose of avoiding a premature convergence in the automatic search, we propose a population restarting process that introduces diversity in the search. On the one hand, this process maintains the elitist generation and, on the other hand, attempts to keep the diversity of the population.

The restarting process is triggered when the threshold maintained by incest prevention operator reaches zero value (the process of threshold updating is explained previously in this same section). This threshold sets a minimum distance that two individuals must overcome in order to be crossover and generate offspring. If during a generation no crossover can be performed because their individuals are very similar, the threshold decreases to be less restrictive the next generation. If the threshold reaches the zero value, the population has converged to a local optimum, so the population is restarted.

Following, it is described the procedure followed by our proposal to carry out the restarting process:

- It is carried out elitism. Thus, the 10% of the best individuals of the pop-

ulation are kept introducing them in the new generation without modifications.

- It is restarted the rest of population. Thus, the rest of individuals to reach the size of population is randomly generated.
- The value of incest prevention threshold ($d$) is restarted and it is set to its initial value.

#### 3.3.3. Fitness function

The fitness measures the goodness of each individual using the Root-Mean-Squared Error (RMSE) between the estimated ratings given by the RS and the real ratings. This measure tends to penalize larger errors more severely than other metrics. If $p_{i,j}$ is the predicted rating for student $i$ over course $j$, and $v_{i,j}$ is the true rating and $K = \{(i, j)\}$ is the set of student-course ratings to predict. Then the RMSE, whose purpose is to minimize, it is defined as:

$$RMSE = \sqrt{\frac{\sum_{(i,j)\in K}(p_{i,j} - v_{i,j})^2}{\#K}} \quad (5)$$

In the calculation of fitness function, it is applied a hold-out process where the 80% of the data are used by the RS for predicting a recommending on the remaining 20%.

### 3.4. Complexity study

This section studies the time complexity of the model proposed. The model is composed of two systems combined in a hybrid model. Thus, the different components are evaluated.

1. For the CF system, our proposal has to process the different students based criteria and carry out the estimation of ratings. Concretely:
    - Processing ratings and grades criteria have a cost of $O(S^2C)$ each one of them, where $S$ is the number of students in the system and $C$ is the number of courses.

- Processing branch criterion has a cost of $O(S^2)$.
- Combining the criteria in the final similarity measure has a cost of $O(S^2)$ and finding the $K$ nearest neighbors of all the students has a cost of $O(S^2 \log(K))$.
- Finally, estimating the unknown ratings has a cost of $O(SCK)$.

2. For the CBF system, our proposal has to process the different course based criteria and carry out the estimation of ratings. Concretely:

   - Processing professors and competences criteria has a cost of $O(C^2P)$ and $O(C^2T)$ respectively, where $P$ is the number of professors and $T$ the number of competences.
   - Processing departments and contents criteria has a cost of $O(C^2)$ each one of them.
   - Combining all the criteria has a cost of $O(C^2)$.
   - Finally, estimating the unknown ratings has a cost of $O(SC^2)$.

3. For our hybrid RS, it is necessary to combine the estimations of the two previous systems (with a cost of $O(SC)$). Thus, summing up all complexities our SR has a cost of $O(S^2(CK + \log(K)) + C^2(P + T + S))$.

Concerning to the time complexity of the GA used to optimize the parameters of the RS, it is determined by the size of the population $N$ and the length of the genotypes $L$, and with it, the fitness function to optimize. Based on these parameters the cost is $O(N(S^2(CK + \log(K)) + C^2(P + T + S)))$. The other operations to take into account are the crossover and mutation operators witha cost of $O(NL)$, respectively. In conclusion, the total complexity time of the GA is $O(N(S^2(CK + \log(K)) + C^2(P + T + S) + L))$.

## 4. Experimental study

The developed RS has been implemented using the Apache Mahout framework [29] for distributed linear algebra and the GA has been developed using the software for evolutionary computation in Java JCLEC [30]. All the experiments have been executed in a machine with Ubuntu 16.04 64 bits operative system, AMD Ryzen 5 1600 processor and 4 GiB of RAM.

Table 2: Configuration of GA parameters

| Parameter | Value |
|---|---|
| Number of generations | 1000 |
| Population size | 50 |
| Crossover probability | 0.9 |
| Initial value for incest prevention threshold | 4 |
| Allowed range for weight genes | [0, 50] |
| Allowed range for neighborhood gen | [1, 50] |
| Allowed range for metrics genes | [0, 4] or [0, 1] |

As mentioned in section 3.1, the dataset used for the evaluation experiments comes from real ratings and grades gathered from students of Computer Science in University of Cordoba. To train the RSs and apply the GA are used training data. Then, test data are used to check the performance of RSs. More detailed are given in section 4.2.1.

The experimental evaluation of this work is divided in three phases. In the first phase, it is studied the relevance of each criterion and parameters obtained by the automatic optimization carried out by the GA. In the second one, it is studied the performance of our proposal compared with the CF model and the CBF model using both multiple criteria and using individual criteria. Also, a comparison with previous works is included. Finally, an example of specific recommendation of courses given by our system to a particular student is analyzed.

### *4.1. Influence of the different criteria in the Recommendation System*

The weight assigned automatically to each criterion determines the influence of different criteria in the recommendation process. In this section, first it is analyzed the best configuration obtained by the GA. Then, an in-depth study of the evolution of the weights and the other parameters is carried out. The parameter configuration of GA is shown in Table 2.

Table 3: Criteria weights and similarity measures chosen by the GA

| **Hybrid SR** | |
|---|---|
| CF weight | 0.54 |
| CBF weight | 0.46 |
| **CF (based in student information)** | |
| Ratings (sim. metric) | Pearson correlation |
| Grades (sim. metric) | Pearson correlation |
| Ratings (weight) | 0.60 |
| Grades (weight) | 0.30 |
| Branch (weight) | 0.10 |
| Neighborhood size | 15 |
| **CBF (based in course information)** | |
| Professors (sim. metric) | Jaccard index |
| Competences (sim. metric) | Jaccard index |
| Professors (weight) | 0.65 |
| Competences (weight) | 0.00 |
| Knowledge area (weight) | 0.00 |
| Contents (weight) | 0.35 |

Table 3 shows the weights assigned to each criterion, as well as, the similarity measures provided by the GA for configuring our hybrid RS. The results bring to light the importance of using a hybrid system that combines both student and course information. Thus, the relevance of these models to obtain the best recommendations is balanced assigning a weight of 0.54 to CF model and 0.46 to CBF model. Concerning to CF based on student information, we see that the most important criterion is the rating criterion (0.60), followed by the grade criterion (0.30). Finally, the branch criterion is the less relevant (0.10). We must also be taken into account a neighborhood size relatively small (15 students). Thus, a course will be recommended to a student mainly whether students with similar ratings and grades rated positively that course. Concerning to CBF based on course information, the most relevant criterion is assigned to the professor criterion (0.65) followed by the content criterion (0.35), while the competences and knowledge area seem to be irrelevant. These results reveal the significance of professors of each course, being a factor slightly more important than even the contents of course. Thus, a course will be recommended to a student if the professors that teach the course and the contents are similar to other courses that were interesting for the student in the past.

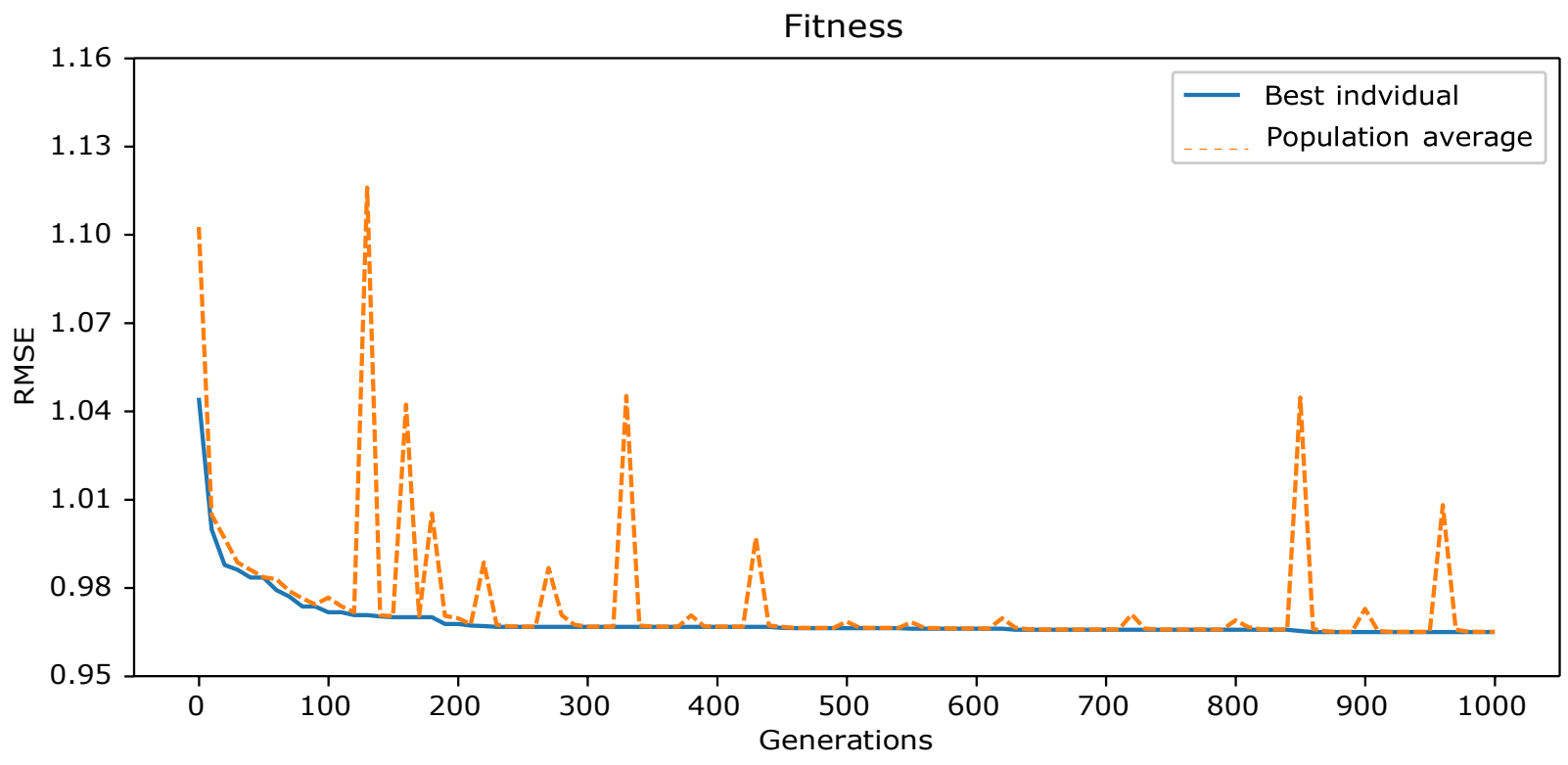

Figure 8: Fitness evolution in genetic search

Concerning to the evolution of the weights of each criterion in the automatic search, as well as the rest of parameters, they are analyzed through the gener- ations of the proposed GA. Figure 8 shows the optimization of the fitness, thatis, the minimization of the RMSE between estimated and real ratings. Herewe can see, on one hand, the effects of restarting the population with the in- crease peaks presented by the average fitness values. On the other hand, the best individuals found is keept in each generartion, so the best individual fitness value never gets worse, although it is increasingly slowly. Finally, we obtain a RMSE around 0.96 in the best case. The evolution of the RS configurations

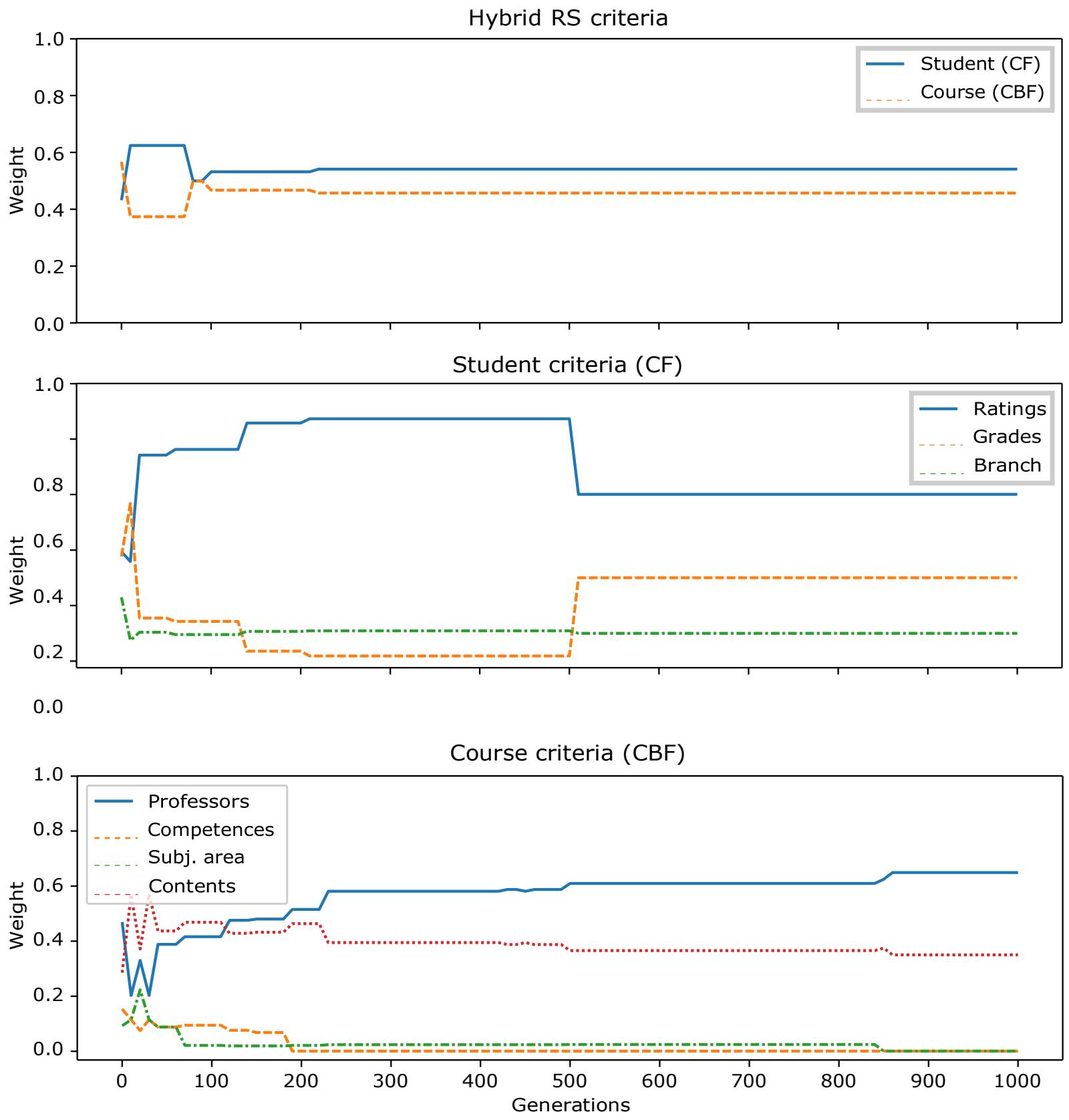

Figure 9: Criteria weight evolution in the GA

that produce this best individual is represented in Figures 9 and 10.

Attending to the importance given to each criterion, Figure 9 shows the evolution of the weights assigned to each criterion and to each system. The main conclusions are:

- Concerning to general weights for the hybrid RS corresponding to student and course information, we can see that are weights very stable and tend to stay balanced from the early generations, although the student information seems to be slightly more important.

- Concerning to evolution of specific student criteria, we can appreciate two phases: in the first half of the experimentation the rating criterion tends to monopolize all the importance, but finally it is balanced with the grades criterion. Even so, rating information appears as the most relevance, followed by grade criterion that also has a considerable importance. On the other hand, the branch criterion does not seem to contribute much to recommendations.

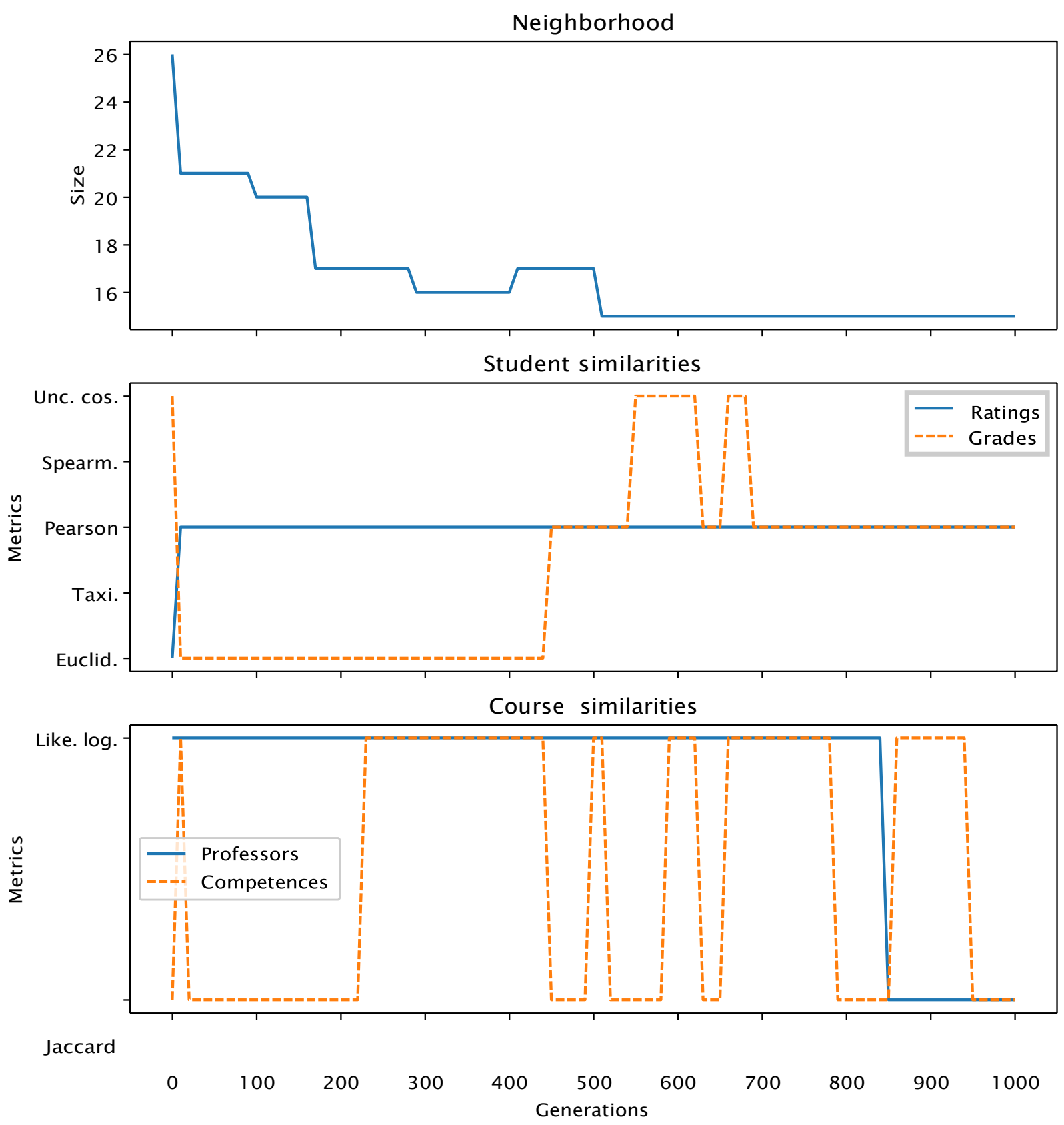

Figure 10: Neighborhood and similarity metrics evolution in the GA

Looking at specific course criteria, we can appreciate more instability, mainly in first generations (it is shown in Figure 9). Even so, there is a clear trend in which professors and contents have the main importance, while competences and knowledge area are practically irrelevant. The low relevance of knowledge area criterion may be due to the characteristic of the data studied: most of the considered courses belong to the same knowledge area, resulting in that this factor does not provide relevant information. On the other hand, the low relevance of competences may be due to they are too generic and many courses share the same.

Attending to the rest of RS configuration, Figure 10 shows the evolution of the best RS configuration on neighborhood and metrics used to carry out the similarities between students and courses. We can see that the size of neighbor- hood is shrinking. About similarity metrics considered in student information, it seems that the Pearson correlation coefficient works better than the rest for both ratings and grades criteria. Even so, in grades we can appreciate more in- stability, highlighting the Euclidean distance, which is a local optimum during the first half of the search. Looking at similarity metrics considered in course information, we can see that for professors, it changes from logarithmic likeli- hood to Jaccard index according to the optimization of weight for this criterion. In the case of similarity measure for competences criterion, it can be seen thatit changes frequently, although it seems that this selection does not affect highly to the final result, due to that, finally, the weight for this criterion approaches the zero value.

### *4.2. Comparison of performance with other models*

In order to study the relevance of our proposal is carried out two studies. On one hand, it is shown the advantages of our proposal compared to the use of CF and CBF independently, as well as, the use of multiple criteria or single criterion to recommend courses. Thus, the relevance of using the most appro- priate criteria and configuration is shown. On the other hand, it is carried out a comparative study with previous proposals to show the good performance of our proposal.

#### *4.2.1. Setting experimental study*

For evaluating the RS is used a cross-validation approach with five folds in order to obtain statistically significant results [31]. The base of this process is to divide the ratings of each student in train and test. Thus, the RS tries to estimate those ratings that have been hidden for test using only the information of training. We propose to follow a stratified approach to build the folds basedon keeping a balance between the number of ratings received per course across the different partitions. The whole process is described in Figure 11.

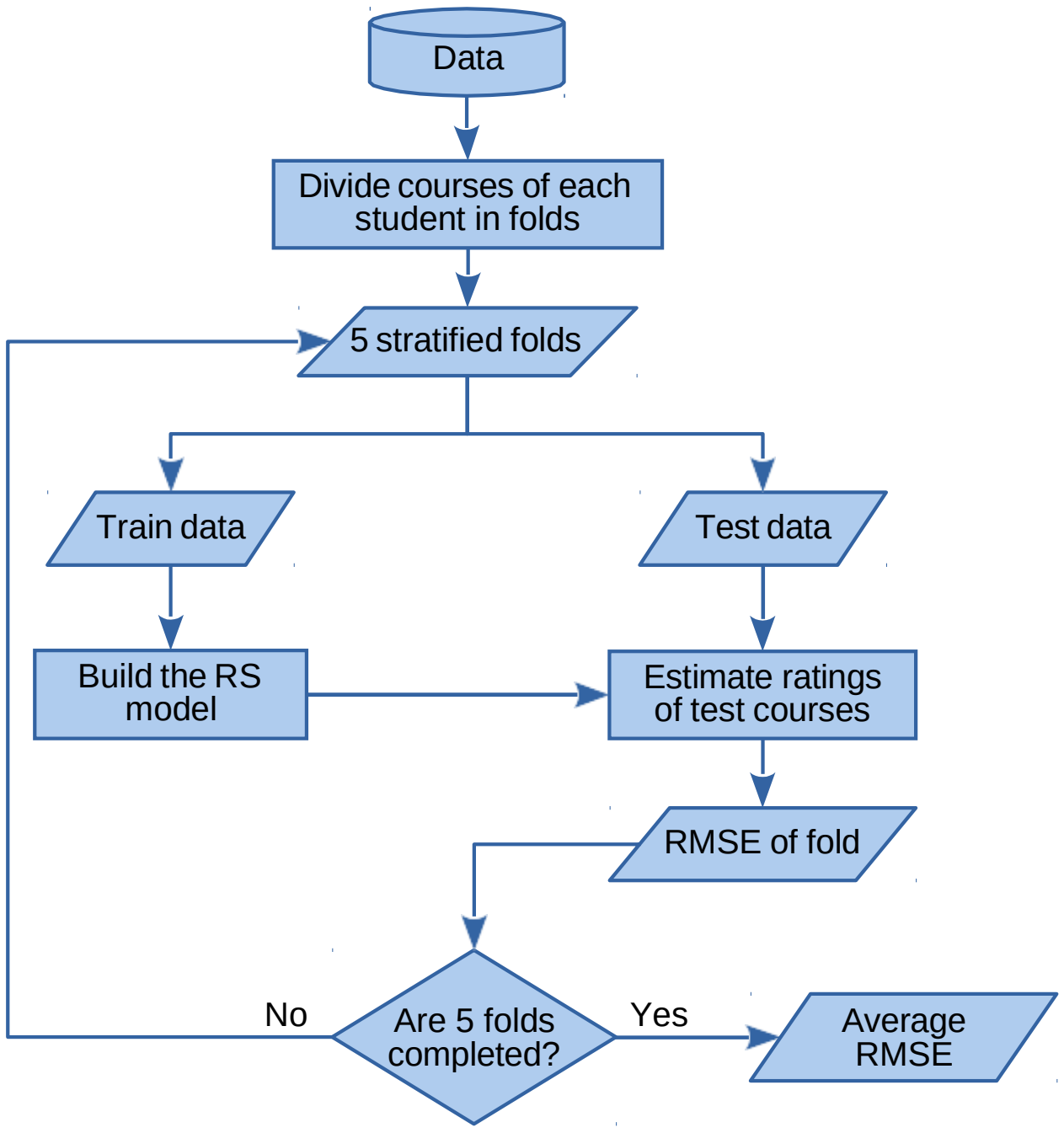

Figure 11: Cross validation carried out in experimental study

With real and estimated ratings, we can compute diverse error metrics to evaluate RSs. Specifically, it is used the following:

- RMSE. It is the root mean squared error between the estimated ratings given by the RS and the real ratings. The definition of this measure can be found in Section 3.3.3).

- nDCG. It is normalized discount cumulative gain. This measure is based on Information Retrieval (IR) techniques. It is not centered in how much differ real and estimated rating, but on how relevant are the recommended courses. In this case, the evaluation process consists of hiding the most relevant ratings for a student and training the RS with the remaining data. After that, the RS is asked to give a recommendation for the given student of so many courses as was hidden. With the ordered lists of real most relevant courses and those estimated, we can compute diverse IR metrics. Specifically, nDCG is related with measuring of ranking quality.

$$nDCG = \frac{DCG}{IDCG} \tag{6}$$

DCG at a particular rank position $p$, if $rel_i$ is the graded relevance of the result at position $i$, is defined as:

$$DCG = \sum_{i=1}^{p} \frac{rel_i}{\log_2(i+1)} \tag{7}$$

Normalization is given by the division by the Ideal DCG at position $p$ (IDCG). If $|REL|$ is the list of relevant courses (ordered by their relevance) in the corpus up to position $p$, the IDCG is defined in equation 8.

$$IDCG = \sum_{i=1}^{|REL|} \frac{2^{rel_i} - 1}{\log_2(i+1)} \tag{8}$$

- Reach. It is evaluated the possibility to carry out a recommendation. CF systems are based on similarities between students. Depending on the criteria used, there are some outlier students for which no satisfactory similarities are found, and so no recommendation can be made for them. In order to measure this behavior we also consider the reach of the RS, whose purpose is to maximize. If $K = \{(i, j)\}$ is the set of student- course ratings to predict and $p_{i,j}$ is the predicted rating, reach is defined in equation 9.

$$Reach = \frac{\#K - \sum_{(i,j)\in K} p_{i,j}}{\#K} \text{ A } p_{i,j} = \emptyset \tag{9}$$

- Execution time. It is also a relevant approach. The mean execution time is analyzed once each model has been learned. It is calculated the time that each approach takes on building the recommendation ranking for a user. As it has been mentioned, the experimentation has been carried out by a machine with Ubuntu 16.04 64 bits as operative system, AMD Ryzen 5 1600 processor and 4 GiB of RAM. The specific execution time of GA is not included in the execution time of our proposal (proposed hybrid RS). GA is applied as a previous step to configure the parameters of hybrid RS. Once these parameters and weights are configured, the RS is carried out according to them, so it is not necessary to execute the GA again. The execution time taken by the GA executed during 500 generations and a population size of 100 individuals to obtain the configuration is 5 hours and 16 minutes.

Table 4: Experimental results of different RS

| Approach | RMSE | nDCG | Reach (%) | Time (s) |
|---|---|---|---|---|
| Proposed hybrid RS | 0.971 | 0.682 | 100.00 | 3.022 |
| CF with multi-criteria | 1.123 | 0.709 | 79.30 | 1.582 |
| CBF with multi-criteria | 1.206 | 0.186 | 100.00 | 1.324 |
| CF with rating criterion | 1.198 | 0.635 | 95.09 | 1.020 |
| CF with grade criterion | 1.347 | 0.534 | 96.14 | 1.014 |
| CF with branch criterion | 1.221 | 0.644 | 88.42 | 0.250 |
| CBF with professor criterion | 2.608 | 0.203 | 100.00 | 0.785 |
| CBF with content criterion | 1.224 | 0.234 | 100.00 | 1.874 |
| CBF with competences criterion | 1.229 | 0.145 | 100.00 | 0.833 |
| CBF with knowledge area criterion | 1.564 | 0.237 | 100.00 | 0.370 |

#### *4.2.2. Evaluating the relevance of our proposal*

In this section, it is evaluated the relevance of using our hybrid model and the specific criteria considered. For this purpose, a comparative study considering the CF and CBF independently and using different criteria has been carried out. The results are shown in Table 4. In the first row, it is represented the proposed hybrid RS results, in the two following rows, the multi-criteria RS results based on CF and based on CBF using all student criteria and all course criteria and with the same configuration given by the GA for the hybrid RS. From the fourth to the sixth row, it is shown the results of RS based on CF using only one criterion in each case. Similarly, from the seventh to the tenth row, it is shown the results of RS based on CBF using only one criterion in each case.

The results prove the relevance of using a hybrid approach with multiple criteria whose estimations are significantly better that the rest of the models (that is, the RMSE values are the lowest). It is more, the use of multiple criteria works better than the mono criterion versions of each one. In general, we can see the information of the students (student criteria) are more helpful to make recommendations, as show the lower RMSE for CF than for CBF. In addition, it is relevant to highlight the low capacity of CBF to offer a list of relevant courses for a student (low nDCG). However, CF could have problems offering recommendations for all users since some of them are too different from rest and the RS can not build a neighborhood (reach value less than 100%). On the other hand, CBF can provide recommendations for all the students. In the hybrid RS, the advantages of CF and CBF are combined, so we get 100% of reach and more accurate recommendations.

It is also worth noting that any criteria that are more relevant than other ones when they are combined, however they obtain worse results when they are used as single criterion. For example, the professor criterion is relevant, but there are few courses taught by the same professors. Then, it is necessary to have other information to be able to recommend new interesting courses for a student. Hence the importance of combining and weighting correctly in this systems all information available.

#### *4.2.3. Comparison of performance with related work*

Due to the data limitation exiting in course recommendation caused by lack of public datasets and the fact that each work uses different criteria to carry out the recommendation, it is not common to find comparisons between proposals.

After analyzing previous proposals, the following works can be included in the comparison: Ma et al. [20] apply clustering based on courses descriptions to make the recommendations, Unelsrod [21] combines user-based CF with item- based filtering using ratings, area of the courses and professors to make the recommendations and the generic proposal of multi-criteria RS presented by Shambour and Lu [32] that it has been implemented because of its similarities with our work of combining the multi-criteria at level of computing of similari- ties. In this case, we have used the competences to build the semantic module and the ratings and grades to build the multi-criteria item-based CF. Table 5 shows the results obtained by the different models. It can be seen that our pro- posal obtains more accuracy estimations and most relevant recommendations, at the expense of a bigger time of response per recommendation.

*4.3. Study case for a specific student*

With the purpose of studying the performance of our RS with an specific case, it has been designed an experiment that shows the courses that our RS would recommend to a specific student. In this experiment, it is selected ran- domly a student of our database. Concretely, the student identified by the ID 14. Then, it is removed of database 8 of his/her courses ratings considering both courses with a high rating and low rating. Finally, it is executed our pro- posed hybrid RS (see configuration in Table 3). After that, our RS provides a predicted raking for the student 14 and each one of the courses previously elim- inated. In this step, it is evaluated the values of real ratings and the estimated ratings to determine if our system would recommend or would not recommend relevant courses. For this study, it is considered that a rating above 2.5, it would recommend the course. However, as our system provides an estimated rating, this assumption could be adapted to other more specific assumptions for particular users.

Table 5: Comparison with related work

| Approach | RMSE | nDCG | Reach (%) | Time (s) |
|---|---|---|---|---|
| Proposed hybrid RS | 0.971 | 0.682 | 100.00 | 3.022 |
| CBF with clustering [20] | 1.224 | 0.234 | 100.00 | 0.057 |
| User-based&item-based CF [21] | 1.166 | 0.549 | 100.00 | 0.271 |
| MCSeCF [32] | 1.595 | 0.112 | 100.00 | 5.371 |

Table 6: Rating estimations for an specific student

| Course ID | Real Rating | Estimated Rating | Relevant Course | Recommended Course |
|---|---|---|---|---|
| 1 | 5.00 | 4.18 | Yes | Yes |
| 6 | 4.00 | 3.91 | Yes | Yes |
| 8 | 1.00 | 1.08 | No | No |
| 13 | 3.00 | 3.37 | Yes | Yes |
| 15 | 3.00 | 2.20 | Yes | No |
| 17 | 4.00 | 4.20 | Yes | Yes |
| 18 | 1.00 | 0.89 | No | No |
| 24 | 4.00 | 1.94 | Yes | No |

The results are showed in Table 6. Attending to the real and estimated ratings, it can be calculated the error in estimations. Thus, for this student the RMSE is 0.8484. Attending to the relevance of recommendations, RS finds 4of 6 relevant courses. However, this type of evaluation can be quite ambiguous, since certain ratings in the middle of the range can have different meanings depending on the user. See for example the case of the courses 13 and 15, rated with a medium value: their respective estimations do not differ by more thanhalf a point from reals, however, one is marked as relevant and the other is not. It has been established that the RS recommend the top 3 courses more relevantfor the student, accompanied by the rating estimation in order to give to the student more information with which he/she can take the final decision. So, attending to the recommendations that student 14 would receive, they would be *[17(4.20), 1(4.18), 6(3.91)]*.

## 5. Conclusions

It is presented a hybrid multi-criteria RS applied to recommendation of university courses. The proposed model combines information of the student and the course using various tools such as CF based on neighborhood and CBF and semantic analysis. We want to emphasis on how this information is combined by mean of configurable weights to determine the relevance of each criterion. In this way, an adapted GA has been implemented that produces understandable models in which we can control the relevance of each criterion in the recommen- dations, as well as obtain the best configuration of all parameters used in the RS, such as similarity measures and number of neighbors. Experimental results show that considering several criteria provides better results, but it is necessary to study the relevance of each of them, since not all factors are equally relevant.In addition, the use of a hybrid system which combines both CF and CBF also optimizes the results achieved.

As future work, we propose the inclusion of constrains to recommendations that help to students to filter courses by semester, academic year and other parameters. Further, we aim to extend the criteria taken into account to more courses of other degrees and obtain more data from students that allow to make more tests and, ultimately, to generalize the obtained conclusions to other educative areas. Another interesting future line research would be the inclusion of social network analysis to handle trust in social networks using reputation mechanism that captures the implicit and explicit connections between the network members to improve the recommendations.

## Acknowledgments

This work was supported by the Spanish Ministry of Science and Technology [No. TIN2017-83445-P].

## References

[1] O. Iatrellis, A. Kameas, P. Fitsilis, Academic advising systems: A system- atic literature review of empirical evidence, Education Sciences 7 (4) (2017)1–17.

[2] F. O. Isinkaye, Y. O. Folajimi, B. A. Ojokoh, Recommendation systems: Principles, methods and evaluation, Egyptian Informatics Journal 16 (3) (2015) 261–273.

[3] A. Al-Badarenah, J. Alsakran, An automated recommender system for course selection, International Journal of Advanced Computer Science and Applications 7 (3) (2016) 166–175.

[4] B. Bakhshinategh, G. Spanakis, O. Zaiane, S. ElAtia, A course recommender system based on graduating attributes, in: Proceedings of the 9th International Conference on Computer Supported Education (CSEDU), Vol. 1, SCITEPRESS, 2017, pp. 347–354.

[5] C. Romero, S. Ventura, Data mining in education, Wiley Interdisciplinary Reviews: Data Mining and Knowledge Discovery 3 (1) (2013) 12–27.

[6] A. Peña-Ayala, Educational data mining: A survey and a data mining-based analysis of recent works, Expert Systems with Applications 41 (4) (2014) 1432–1462.

[7] J. Bobadilla, F. Ortega, A. Hernando, A. Gutiérrez, Recommender systems survey, Knowledge-Based Systems 46 (2013) 109–132.

[8] J. Lu, A Personalized e-Learning Material Recommender System, in: 2nd International Conference on Information Technology and Applications (ICITA 2004), 2004, pp. 374–380.

[9] J. Miranda, P. A. Rey, J. M. Robles, Udpskeduler: A web architecture based decision support system for course and classroom scheduling, Decision Support Systems 52 (2) (2012) 505–513.

[10] A. Mohamed, A decision support model for long-term course planning, Decision Support Systems 74 (2015) 33–45.

[11] G. Linqi, L. Congdong, Hybrid personalized recommended model based on genetic algorithm, in: International Conference on Wireless Communications, Networking and Mobile Computing (WiCOM 2008), 2008, pp. 1–4.

[12] C.-S. Hwang, Genetic algorithms for feature weighting in multi-criteria recommender systems, Journal of Convergence Information Technology 5 (8) (2010) 126–136.

[13] J. Bobadilla, F. Ortega, A. Hernando, J. Alcalá, Improving collaborative filtering recommender system results and performance using genetic algorithms, Knowledge-Based Systems 24 (8) (2011) 1310–1316.

[14] M. Salehi, M. Pourzaferani, S. A. Razavi, Hybrid attribute-based recommender system for learning material using genetic algorithm and a multidimensional information model, Egyptian Informatics Journal 14 (1) (2013) 67–78.

[15] K. Taha, Automatic academic advisor, in: 8th International Conference on Collaborative Computing: Networking, Applications and Worksharing (CollaborateCom), IEEE, 2012, pp. 262–268.

[16] K. Ganeshan, X. Li, An intelligent student advising system using collaborative filtering, in: Proceedings - Frontiers in Education Conference (FIE), IEEE, 2015, pp. 1–8.

[17] P. C. Chang, C. H. Lin, M. H. Chen, A hybrid course recommendation system by integrating collaborative filtering and artificial immune systems, Algorithms 9 (3).

[18] L. Mostafa, G. Oately, N. Khalifa, W. Rabie, A case based reasoning system for academic advising in egyptian educational institutions lamiaa, in: 2nd

International Conference on Research in Science, Engineering and Technology (ICRSET'2014),, 2014, pp. 5–10.

[19] C. Y. Huang, R. C. Chen, L. S. Chen, Course-recommendation system based on ontology, in: International Conference on Machine Learning and Cybernetics (ICMLC), Vol. 3, IEEE, 2013, pp. 1168–1173.

[20] H. Ma, X. Wang, J. Hou, Y. Lu, Course recommendation based on semantic similarity analysis, in: 3rd International Conference on Control Science and Systems Engineering (ICCSSE), IEEE, 2017, pp. 638–641.

[21] H. F. Unelsrød, Design and evaluation of a recommender system for course selection, Ph.D. thesis, Norwegian University of Science and Technology (2011).

[22] O. Daramola, O. Emebo, I. Afolabi, C. Ayo, Implementation of an intelligent course advisory expert system, International Journal of Advanced Research in Artificial Intelligence 3 (5) (2014) 6–12.

[23] Z. Gulzar, A. A. Leema, G. Deepak, Pcrs: Personalized course recom-mender system based on hybrid approach, Procedia Computer Science 125 (The 6th International Conference on Smart Computing and Communications) (2018-01) 518–524.

[24] D. Wu, J. Lu, G. Zhang, A Fuzzy Tree Matching-Based Personalized E-Learning Recommender System, IEEE Transactions on Fuzzy Systems23 (6) (2015) 2412–2426.

[25] L. J. Eshelman, The chc adaptive search algorithm: How to have safe search when engaging in nontraditional genetic recombination, Foundations of genetic algorithms 1 (1991-01) 265–283.

[26] J. Derrac, S. García, F. Herrera, A first study on the use of coevolutionary algorithms for instance and feature selection, in: International Conference on Hybrid Artificial Intelligence Systems, Springer Berlin Heidelberg, 2009, pp. 557–564.

[27] D. Peralta, S. del Río, S. Ramírez-Gallego, I. Triguero, J. M. Benitez, F. Herrera, Evolutionary feature selection for big data classification: A mapreduce approach, Mathematical Problems in Engineering 2015 (2015)
1–12.

[28] R. W. Hamming, Coding and Theory, Prentice-Hall Englewood Cliffs, 1980.

[29] E. Friedman, T. Dunning, R. Anil, S. Owen, Mahout in Action, ManningPublication, 2011.